\documentclass[lettersize,journal]{IEEEtran}
\usepackage{amsmath,amsfonts}
\usepackage{algorithmic}
\usepackage{algorithm}
\usepackage{array}
\usepackage[caption=false,font=normalsize,labelfont=sf,textfont=sf]{subfig}
\usepackage{textcomp}
\usepackage{stfloats}
\usepackage{url}
\usepackage{verbatim}
\usepackage{graphicx}
\usepackage{cite}
\usepackage{amssymb}
\usepackage{amsmath}
\usepackage{booktabs}
\usepackage{multirow,pifont}
\usepackage{xcolor}

\usepackage[colorlinks=true, linkcolor=blue, citecolor=blue, urlcolor=blue, breaklinks=true]{hyperref}
\definecolor{orangeyellow}{rgb}{1.0, 0.68, 0.0}  
\definecolor{darkred}{rgb}{0.55, 0.0, 0.0}       
\definecolor{custompurple}{rgb}{0.63, 0.26, 0.94}

\graphicspath{{figures/}}

\begin{document}

\title{VesselBench-800K: A Large-scale Perception Benchmark for Multimodal Vessel Detection, Counting, and Density Estimation}

\author{Danfeng Hong,~\IEEEmembership{Senior Member,~IEEE,}
        Chenyu Li,
        and Jocelyn Chanussot,~\IEEEmembership{Fellow,~IEEE}

\thanks{This work was supported by the National Natural Science Foundation of China under Grant 42241109 and Grant 42271350, by the International Partnership Program of the Chinese Academy of Sciences under Grant No.313GJHZ2023066FN.}
\thanks{D. Hong is with the School of Automation, Southeast University, Nanjing 211189, China. (e-mail: danfeng.hong@seu.edu.cn)}
\thanks{C. Li is with Univ. Grenoble Alpes, INRAE, LESSEM, 38000 Grenoble, France. (e-mail: chenyuli.erh@gmail.com)}
\thanks{J. Chanussot is with Univ. Grenoble Alpes, Inria, CNRS, Grenoble INP, LJK, Grenoble 38000, France. (e-mail: jocelyn.chanussot@grenoble-inp.fr)}
}

\markboth{}%
{Shell \MakeLowercase{\textit{et al.}}: A Sample Article Using IEEEtran.cls for IEEE Journals}


\maketitle

\begin{abstract}
Vessel perception from space is crucial for a wide range of maritime applications, from traffic monitoring to environmental protection. However, most existing datasets predominantly focus on general object detection tasks in optical remote sensing (RS) images. Relying solely on single-modality optical RS images proves inadequate for effectively perceiving vessel objects in complex maritime scenarios, where ever-changing weather conditions (e.g., clouds and rain), the need for day-and-night coverage, and the inherent limitations of a single imaging modality pose significant challenges. To fill this gap, we introduce VesselBench-800K, the largest-to-date benchmark dataset on a global scale for vessel perception in multimodal RS images. As its name suggests, VesselBench-800K comprises 800,000 images, each at a resolution of 512$\times$512 pixels, specifically curated for vessel perception tasks such as detection, counting, and density estimation. These multimodal image pairs (i.e., optical, SAR) are collected from diverse platforms, sensors, scenes, shooting heights, and synthetic sources, spanning spatial resolutions from 4.5m to 0.1m. Furthermore, we evaluate numerous state-of-the-art detection, counting, and density estimation models on VesselBench-800K through both qualitative and quantitative comparisons. By revealing previously unrecognized cues, this dataset holds immense potential to significantly advance our understanding of marine traffic. Our VesselBench dataset will be publicly available at \url{https://github.com/danfenghong/IEEE_TGRS_VesselBench} to support and contribute to community development.
\end{abstract}

\begin{IEEEkeywords}
Benchmark, Counting, Density estimation, Detection, Large-scale, Multimodal perception, Vessel.
\end{IEEEkeywords}

\section{Introduction}
\IEEEPARstart{M}{aritime}  traffic is essential for global trade, security, and ecological stewardship, necessitating advanced monitoring and perception systems \cite{paolo2024satellite}. Accurate vessel detection and recognition enhance maritime safety and protect ecosystems, providing essential insights into ship routes, traffic density, and potential hazards. This allows stakeholders from various sectors to make decisions that balance economic and ecological priorities. Despite the importance, the vastness and remoteness of maritime areas complicate data collection, often leading to incomplete vessel activity mappings. The critical need to understand human impacts on marine environments, compounded by the challenges of climate change and increasing human activities, drives the pursuit of more effective maritime monitoring strategies. Extensive efforts focus on developing robust systems that ensure sustainable ocean management and contribute to the long-term health of marine ecosystems \cite{march2021tracking,womersley2022global}.

Recent advancements in Earth observation (EO) and remote sensing (RS) technologies, along with the deployment of numerous satellite constellations, have dramatically increased the volume of observational data available. This surge in data provides a critical opportunity for enhanced perception of vessels in large-scale and even global marine environments. Over the past decades, vessel detection and recognition in RS images relied on manually designed features, such as texture, statistics, shape, and structure. However, the advent of deep learning (DL) \cite{lecun2015deep} has revolutionized this field. Contemporary state-of-the-art (SOTA) methodologies that leverage DL have significantly advanced object detection capabilities in RS images \cite{xia2018dota,xie2021oriented,li2023large,han2021redet,ding2019learning,luo2024pointobb}. Despite these advancements, performance gains and practical applications have been suffering from a bottleneck due to the absence of a large-scale and high-quality benchmark with a focus on maritime vessels. We now turn to elaborate on three critical aspects. First, current datasets predominantly focus on local or regional scenes. As a result, the capabilities required for perception and understanding in large-scale and even global scenarios remain largely unexplored and uncertain. Second, existing datasets largely cater to general object detection tasks within unimodal (e.g., optical) RS images. Sole reliance on single-modality RS data is insufficient for effectively perceiving complex scenarios. Dynamic factors such as fluctuating weather conditions (e.g., clouds and rain), the necessity for day-and-night coverage, and the limitations inherent to single-modality imaging pose significant challenges, especially in the unpredictable and ever-changing maritime environment. {Moreover, vessels in real-world maritime scenes exhibit substantial variations in target size and image resolution, ranging from tiny vessels occupying only a few pixels to relatively large targets, which further increases the difficulty of robust vessel perception.} Lastly, much of the current research on RS perception is devoted to terrestrial targets, such as buildings, vehicles, airplanes, water bodies, roads, and trees, with marine traffic receiving considerably less attention. This limits the applicability and scalability of these perception algorithms to complex maritime scenarios.

\begin{figure*}[!t]
   \centering
	\includegraphics[width=1.0\textwidth]{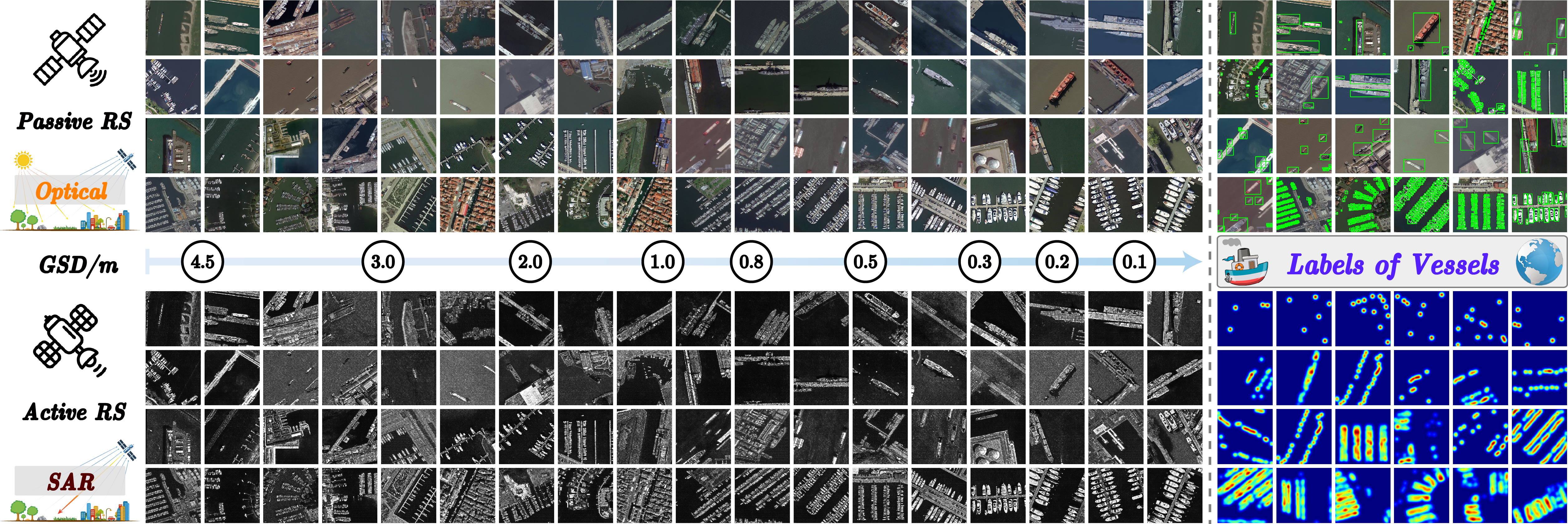}
    \caption{We introduce \textbf{VesselBench-800K}, a large-scale multimodal (passive \textbf{\textcolor{orange}{Optical}} and active \textbf{\textcolor{darkred}{SAR}}) remote sensing (RS) dataset for vessel perception, collected globally. VesselBench-800K features \textbf{800,000} images of 512$\times$512 pixels, with ground sampling distance (GSD) ranging from 4.5m to 0.1m. Vessels are labeled in a horizontal bounding box (HBB) fashion, highlighted in \textbf{\textcolor{green}{green}}, and ground truth (GT) maps for density estimation are generated by means of Gaussian kernel functions. VesselBench-800K caters to three main tasks: detection, counting, and density estimation, making it a comprehensive tool for vessel perception studies.}
\label{fig:datashow}
\end{figure*}

This discussion highlights the necessity for a comprehensive benchmark in maritime vessel perception. Recognizing that single-modality RS data cannot fulfill all-weather and all-day demands, the proposed benchmark, VesselBench-800K, will incorporate multiple modalities, including optical and SAR data, to address these gaps. Additionally, due to its relative inaccessibility compared to land, the challenge of data collection over the sea further complicates the acquisition of diverse and extensive datasets. VesselBench-800K aims to overcome these obstacles, thereby enhancing the capability for large-scale vessel perception. Fig. \ref{fig:datashow} shows the sample examples of multimodal (optical and SAR) RS data in VesselBench-800K, showcasing various GSDs along with HBB detection labels and GTs for density estimation. The main contributions of this paper are outlined as follows:

\begin{itemize}
    \item We created a multimodal RS benchmark dataset, named VesselBench-800K, for marine vessel perception. As indicated by its name, VesselBench-800K consists of 800,000 images, each 512$\times$512 pixels, across optical and SAR modalities. These images collectively feature 1,362,662 vessel instances distributed globally. To the best of our knowledge, this is the first and largest benchmark dataset to date specifically designed for the perception of marine vessels.
    \item VesselBench-800K is collected with diverse spatial resolutions (ranging from 0.1m to 4.5m), platforms (spaceborne and airborne), sensors (optical and SAR), scenes, shooting height, and synthetic sources, covering various marine environments such as offshore, open sea, streams, and rivers.
    \item We evaluate perception performance on VesselBench-800K in three main tasks: detection, counting, and density estimation, where seventeen SOTA algorithms are selected for qualitative and quantitative comparisons in unimodal/multimodal scenarios.
\end{itemize}

\begin{table*}[!t]
    \centering
    \caption{Quantitative comparisons between our \textbf{VesselBench-800K} and existing vessel detection/perception datasets from unimodal (optical, SAR) and multimodal perspectives in terms of nine indices: the number of images (\# of images), image size (pixels), GSD (resolution), number of vessels (\# instances), types of annotations (Annot.), diverse backgrounds (Div.), types of tasks (Det: detection, C: counting, E: density estimation), modality, and data sources, respectively.}
    \resizebox{1\textwidth}{!}{
    \begin{tabular}{c|c|c|c|c|c|c|c|c|c}
        \toprule[1.5pt]
        Datasets & \# of images & Image size (px.) & GSD (m) & \# Instances & Annot. & Div. & Tasks & Modality & Data sources\\
        \hline
        \addlinespace[2pt]
        \multicolumn{10}{c}{\textbf{\textcolor{orange}{Optical}}: Vessel subset in multi-class object detection datasets}\\
        \addlinespace[2pt]
        \hline
        \addlinespace[1.5pt]
        LEVIR \cite{zou2017random} & 22,000 & 800$\times$600 & 0.2$\sim$1.0 & 3,025 & HBB & \ding{52} & Det & RGB & GE (Google Earth)\\
        DIOR \cite{li2020object} & 23,463 & 800$\times$800 & 0.5$\sim$30 & 62,400 & OBB & \ding{52} & Det & RGB & GE\\
        NWPU VHR-10 \cite{cheng2016learning} & 800 & 497$\times$693$\sim$606$\times$1,100 & 0.08$\sim$2.0 & -- & HBB & \ding{56} & Det & RGB & GE, Vaihingen\\
        \cline{10-10} 
        DOTA-v1.0 \cite{xia2018dota} & 2,806 & 800$\times$800$\sim$4,000$\times$4,000 & 0.5 & 20,869 & OBB & \ding{56} & Det & RGB &  \multirow{2}{*}{GE, JL-1, GF-2, Aerial}\\
       DOTA-v2.0 \cite{ding2021object} & 11,268 & 800$\times$800$\sim$20,000$\times$20,000 & 0.5 & 42,517 & OBB & \ding{56} & Det & RGB &  \\
       \cline{10-10} 
       HRRSD \cite{zhang2019hierarchical} & 21,761 & 152$\times$152$\sim$10,569$\times$10,569 & 0.15$\sim$1.2& 4,000 & HBB & \ding{56} & Det & RGB & GE, Baidu Earth\\
        \hline
        \addlinespace[2pt]
        \multicolumn{10}{c}{\textbf{\textcolor{orange}{Optical}}: Dedicated vessel detection datasets}\\
        \addlinespace[2pt]
        \hline
        \addlinespace[1.5pt]
        HRSC2016 \cite{liu2016ship} & 1,680 & 300$\times$300$\sim$1,500$\times$900 & 0.4$\sim$2.0 & 2,976 & OBB & \ding{52} & Det & RGB & GE\\
        LEVIR-Ship \cite{chen2022degraded} & 3,896 & 512$\times$512 & 16 & 3,219 & HBB & \ding{56} & Det & RGB & GF-1, GF-6\\
        VHRShips \cite{kizilkaya2022vhrships} & 6,312 & 1280$\times$720 & -- & 11,337 & HBB & -- & Det & RGB & GE\\
        Airbus \cite{airbus2018airbus} & 192,556 & 768$\times$768 & -- & 213,723 & HBB & \ding{56} & Det & RGB & SPOT-6, SPOT-7\\
        ShipRSImage \cite{zhang2021shiprsimagenet} & 3,435 & 930$\times$930$\sim$1,400$\times$1,000 & 0.12$\sim$6.0 & 17,573 & OBB & \ding{52} & Det & RGB & GE, GF-2, JL-1\\
        \hline
        \addlinespace[2pt]
        \multicolumn{10}{c}{\textbf{\textcolor{darkred}{SAR}}: Vessel (incl.) detection datasets}\\
        \addlinespace[2pt]
        \hline
        \addlinespace[1.5pt]
        AIR-SARShip \cite{xian2019air} & 31 & 3,000$\times$3,000 & 1.0$\sim$3.0 & 461 & HBB & \ding{52} & Det & SAR & GF-3\\
        HRSID \cite{wei2020hrsid} & 5,604 & 800$\times$800 & 0.5$\sim$3.0 & 16,951 & HBB & \ding{52} & Det & SAR & Sen-1B, TerraSAR-X, TanDEM-X\\
        ShipDataset \cite{wang2019sar} & 39,729 & 256$\times$256 & 3.0$\sim$25 & 16,951 & HBB & \ding{52} & Det & SAR & Sen-1, GF-3\\
        FUSAR-Ship \cite{hou2020fusar} & 16,144 & 512$\times$512 & 3.0 & -- & HBB & \ding{52} & Det & SAR & GF-3\\
        SSDD / SSDD+ \cite{zhang2021sar} & 1,160 & 500$\times$500 & 1.0$\sim$15 & 2,456 & HBB / OBB & \ding{52} & Det & SAR & Sen-1, RadarSat-2, TerraSAR-X\\
        OGSOD \cite{wang2023category} & 18,331 & 256$\times$256 & 3.0 & -- & HBB / OBB & \ding{52} & Det & SAR & GF-3\\
        \hline
        \addlinespace[2pt]
        \multicolumn{10}{c}{\textbf{\textcolor{custompurple}{Multimodal}}: Vessel (incl.) detection datasets}\\
        \addlinespace[2pt]
        \hline
        \addlinespace[1.5pt]
         M$^{2}$SODAI \cite{jang2023m} & 1,257 & 224$\times$224, 1,600$\times$1,600 & 0.1$\sim$0.7 & $\sim$ 5,000 & HBB & \ding{52} & Det & RGB, HSI & Aerial (DMC, AsiaFENIX)\\
         \hline
         \addlinespace[1.5pt]
         \textbf{VesselBench-800K (Ours)} & \textbf{800,000} & 512$\times$512 & \textbf{0.1$\sim$4.5} & \textbf{1,362,662} & HBB & \ding{52} & \textbf{Det, C, E} & \textbf{RGB, SAR} & GE, GJ-1, JL-1, GF-2, Aerial, GF-3 \\
        \bottomrule[1.5pt]
    \end{tabular}
    }
    \label{tab:datacompare}
\end{table*}

\section{Related Works}
There has been a general consensus that building very large-scale datasets is not only indispensable and urgent but also represents a strategic advantage in the current era of artificial intelligence (AI). This principle is equally applicable to the fields of EO and RS, where such datasets are crucial for advancing AI technologies to tackle complex challenges in Earth applications, such as maritime traffic and environmental monitoring, in our case. 

To benchmark against renowned datasets in computer vision, e.g., ImageNet \cite{deng2009imagenet}, COCO \cite{lin2014microsoft}, the construction of RS datasets specifically designed for various perception tasks has significantly increased in recent years. Most of these datasets \cite{zou2017random,li2020object,xia2018dota,ding2021object,cheng2016learning,zhang2019hierarchical} consist exclusively of unimodal optical RS images for general multi-class object detection tasks. Following this, a growing number of class-specific object detection datasets are further developed with a particular focus on vehicles, airplanes, buildings, roads, etc. \cite{mundhenk2016large,bastani2018roadtracer,shermeyer2021rareplanes,wang2022learning}. Their creation, to a great extent, promotes the advancement of research and practical applications in these specific object detection categories. It is noteworthy, however, that vessels have received relatively less attention. This is mainly because data collection in marine environments presents more challenges than gathering data for terrestrial targets. While enormous efforts have been made by researchers to collect and build dedicated vessel detection datasets \cite{chen2022degraded,kizilkaya2022vhrships,airbus2018airbus,zhang2021shiprsimagenet,liu2016ship}, these resources remain too small and localized to meet the growing demand for large-scale, globally applicable data. 

Another significant reason for the limited development in vessel detection is the reliance on single-source RS data, specifically optical images. Optical imaging \cite{schaepman2006reflectance}, as a typically passive RS method, is inherently vulnerable to a variety of environmental factors such as cloudy and rainy weather, day-and-night cycles, and fluctuating light intensities. This susceptibility and dependency hamper the reliability, consistency, and robustness of datasets, thereby limiting their effectiveness in diverse maritime conditions. Active radar imaging \cite{richards2009remote} can provide a viable alternative to overcome these environmental limitations. SAR data, which are less affected by weather conditions and lighting, have increasingly been utilized to build datasets for object detection \cite{xian2019air,wei2020hrsid,wang2019sar,zhang2021sar,wang2023category,hou2020fusar,li2024sardet100k}. This use is instrumental in enhancing the scope and accuracy of vessel detection efforts, providing reliable results regardless of environmental changes.

Nevertheless, SAR images are often characterized by lower quality, higher noise levels, and limited identifying information. This inevitably leads to reduced detection and recognition performance. To enhance perception capabilities more effectively and comprehensively, researchers have gradually recognized the need for and started to establish more diverse object detection datasets towards multimodal RS, such as VEDAI \cite{razakarivony2016vehicle} and DroneVehicle \cite{sun2022drone} for vehicle detection, DLR-4K-SAI-LCS \cite{wu2024multimodal} for vehicles and tents, LLVIP \cite{jia2021llvip} for person detection, SeaDronesSee \cite{varga2022seadronessee} for human detection in open water, and M$^2$SODAI \cite{jang2023m} for detecting ships and floating matters. However, these datasets still mainly focus on land targets with RGB + infrared (IR) or DEM or hyperspectral modalities and are generally limited in scale. More importantly, these modalities (\textit{cf.} SAR) have been proven to be ineffective for maritime environmental perception due to the unique challenges posed by the sea, such as water reflections, varying lighting conditions, and the dynamic nature of maritime environments.

{More recently, advances in AI have further promoted multimodal perception and foundation models in RS \cite{hong2026foundation}. By leveraging large-scale pretraining and heterogeneous observations, these studies have demonstrated promising capabilities in learning transferable and generalizable representations across different sensors, modalities, and downstream tasks \cite{li2026fleximo,li2026seamo,li2026any,yin2026disastertd}. Such developments further highlight the importance of large-scale and diverse multimodal benchmarks for advancing general-purpose RS perception.}

Moreover, nearly all existing RS datasets are merely concerned with object detection and often overlook higher-level perceptual needs, such as object counting and density estimation. Exceptions are rare, with only very few works, such as those focused on vehicle perception in \cite{zhao2024vehicle}, addressing these more complex tasks.

Table \ref{tab:datacompare} shows a quantitative comparison between our VesselBench-800K and existing vessel perception datasets, considering several key indices: the number of images (scale), image size, GSD (resolution), number of vessels (instances), types of annotations, types of tasks, diversity of backgrounds, and data sources. As quantified in Table \ref{tab:datacompare}, our VesselBench-800K contains more than ten times the number of images in comparison with other datasets, with a finer and broader range of GSD. These images cover millions of instances across heterogeneous modalities and diverse data sources. The key distinction of VesselBench-800K, in contrast to other datasets, lies in its focus on vessel perception tasks. It goes beyond basic vessel detection to include more comprehensive tasks, such as counting and density estimation.

{Beyond conventional vessel detection, object counting and density estimation have also received increasing attention in RS and computer vision. Recent studies have investigated domain-generalized localization under distribution shifts \cite{gao2025dynamic} and large-scale remote sensing object counting with density-map-based benchmarking \cite{gao2020counting}, highlighting the importance of model robustness and standardized evaluation across diverse scenes. Meanwhile, recent multimodal and open-world counting studies, such as UNICBench \cite{rong2026unicbench}, CountGD \cite{amini2024countgd}, and zero-shot object counting with enhanced quantitative and spatial awareness \cite{zhang2026boosting}, have further extended counting toward multimodal, open-vocabulary, and zero-shot settings. These advances demonstrate an emerging trend from category-specific counting toward more generalizable and multimodal object perception. In this context, VesselBench-800K complements existing studies by providing a large-scale maritime RS benchmark that jointly supports vessel detection, counting, and density estimation across optical and SAR observations, while also providing a foundation for future research on cross-region, cross-time, long-tail, and open-vocabulary maritime perception.}

\section{VesselBench-800K Dataset}

\subsection{Data Acquisition}
The creation of large-scale datasets for vessel perception is a slow, gradual, and long-term process. Unlike land monitoring and perception, many maritime areas are difficult to access and record, leading to operational challenges and adding further burdens to the data collection process. To address this, our strategy involves selectively aggregating open-source RS datasets related to water or vessel tasks. Although each dataset may be small on its own, we follow the principle that ``many a little makes a mickle'' and work to expand their scale to enhance the diversity and applicability of the overall dataset. More specifically, our VesselBench-800K dataset mainly draws from the following six key sources: 1) a subset of DOTA datasets \cite{xia2018dota,ding2021object}; 2) a subset of GLH-water dataset \cite{li2024glh}; 3) a subset of GLH-Bridge dataset \cite{li2024learning}; 4) a subset of QXS-SAROPT dataset \cite{huang2021qxs}; 5) newly-downloaded data from platforms (e.g., Google Earth) and synthetic data. Table \ref{tab:dataacquisition} lists a detailed quantification of the five components that make up VesselBench-800K, including the number of images selected from existing contests and datasets, resampled image size, involved modalities, and labeling details.

{\textbf{Dataset Documentation, Ethics, and Copyright Statement.} Detailed dataset information, including data sources, modalities, spatial resolutions, annotation status, and geographic distributions, is summarized in Table \ref{tab:dataacquisition} and Fig. \ref{fig:distribution}, providing a dataset-card-style overview of VesselBench-800K. The VesselBench-800K dataset is compiled from three primary sources: (1) publicly available open-source datasets, (2) satellite images obtained through manually accessible platforms (e.g., Google Earth), and (3) synthetic data generated using generative models (e.g., CycleGAN in our case). All data usage strictly complies with the respective terms of service of each source. No proprietary or sensitive information is included. The dataset is provided solely for non-commercial, academic research purposes, and is intended exclusively for scientific study and benchmarking, without redistributing any original copyrighted material. It is worth noting that, since VesselBench-800K integrates multiple existing public datasets together with newly collected and generated data, a single unified acquisition time span is not applicable to all images.}

\begin{figure*}[!t]
      \centering
	   \includegraphics[width=1.0\textwidth]{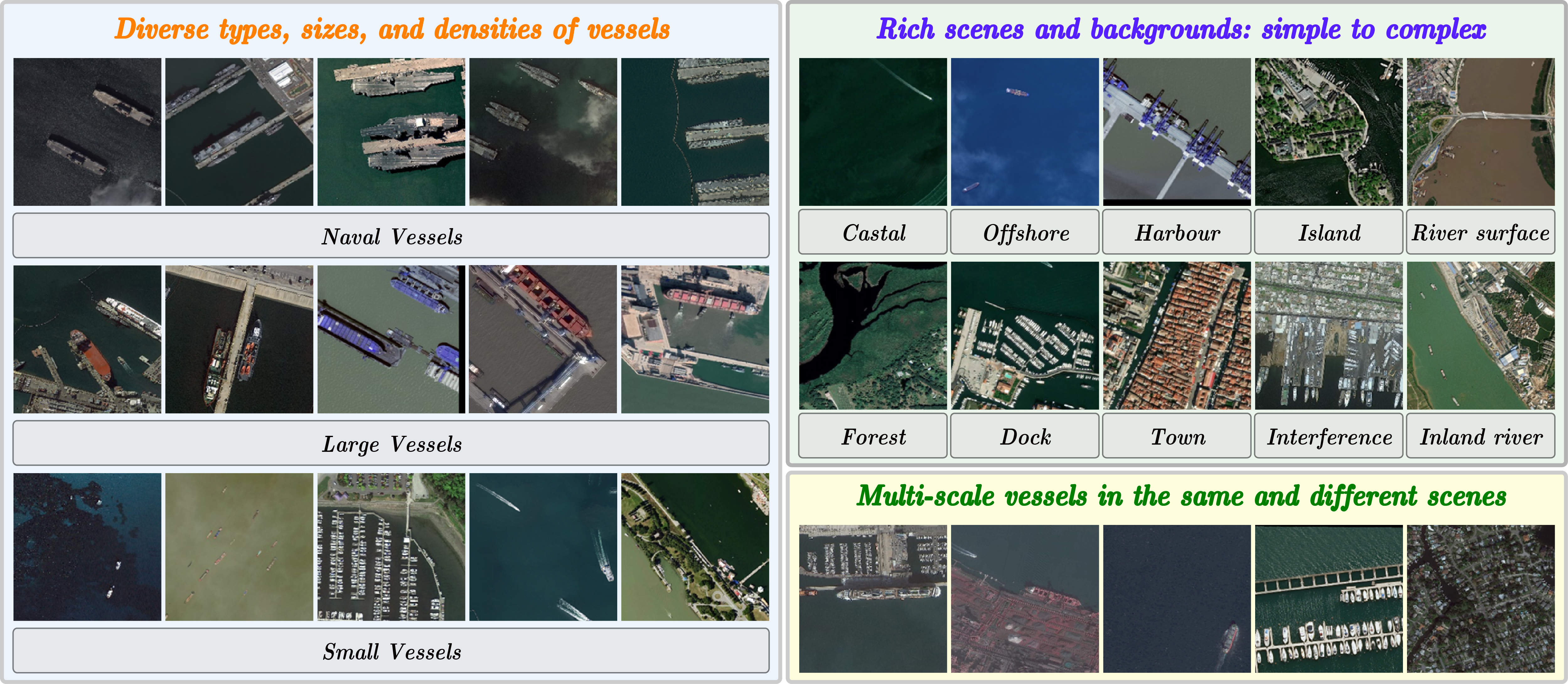}
      \caption{Typical examples from VesselBench-800K to illustrate key data characteristics, including diverse vessel types, sizes, and densities, varied scenarios (ranging from simple to complex), and multi-scale vessels within the same and different scenes.}
\label{fig:samples}
\end{figure*} 

\begin{table}[!t]
    \centering
    \caption{Detailed composition of \textbf{VesselBench-800K}: the number of images (of \# images), involved modalities, GSD, and labeling status (re-w/: re-labeled, w/o: without labels).}
    \resizebox{0.47\textwidth}{!}{
    \begin{tabular}{c|c|c|c|c}
        \toprule[1.5pt]
        Sources & \# of images & Modality & GSD (m) & Labels\\
        \hline
        \addlinespace[1.5pt]
        DOTA & $\sim$22,000 & RGB & 0.5 & re-w/\\
        GLH-water & $\sim$400 & RGB & 0.3 & w/o\\
        GLH-Bridge & $\sim$80,000 & RGB & 0.3$\sim$ 1.0 & w/o\\
        QXS-SAROPT & $\sim$230 & RGB, SAR & 1.0 & w/o\\
        New data  & $\sim$697,370 & RGB, SAR & 0.1$\sim$ 4.5 & w/o\\
        \bottomrule[1.5pt]
    \end{tabular}
    }
    \label{tab:dataacquisition}
\end{table}

\subsection{Data Processing}
\noindent \textbf{Resampling.} Compared to natural images, RS images typically exhibit a wider range of scales (or resolutions), a characteristic that is particularly evident in our case. We, therefore, preprocess these collected, captured, and generated RS images by cropping and resampling them to a unified size of 512$\times$512 pixels. This choice deviates from the commonly used 1024$\times$1024 pixel format in computer vision but ensures consistency across different modalities and data sources while preserving essential spatial information and making it more friendly to extract features of vessel targets at various scales.

\vspace{2pt}
\noindent \textbf{Generation.} Image alignment between modalities plays a crucial role in multimodal tasks, particularly target detection, as it ensures that the spatial correspondence between different data sources is maintained. However, achieving precise alignment is often challenging in reality due to differences in perspective, resolution, the inherent characteristics of each modality, and imaging devices and conditions. To this end, we largely generate synthetic SAR images\footnote{The reason for not generating synthetic optical images from SAR ones is that SAR images tend to be low-quality, highly noisy, have scattering characteristics, and require complex preprocessing. These factors make it challenging to generate high-quality optical counterparts from SAR.} from the corresponding high-resolution and high-quality optical RGB images using CycleGAN \cite{zhu2017unpaired} trained on 20,000 optical-SAR pairs of the QXS-SAROPT dataset. Despite minimal vessel sample pairs available, CycleGAN effectively leverages unpaired samples to bridge the gap between heterogeneous modalities while maintaining spatial consistency. {For multimodal optical-SAR data, spatial correspondence is maintained based on the original paired data when available. For synthetic SAR data, SAR images are directly generated from the corresponding optical images, thereby preserving the spatial locations of vessel targets.} This enables more reliable generation of synthetic SAR data for vessel perception tasks. The well-trained CycleGAN model can then be directly applied to infer synthetic SAR images from input optical RS data. {To ensure the quality of the generated SAR images, we further perform quality checking in terms of visual quality and spatial consistency, with particular attention to the preservation of vessel targets and their spatial locations. The generated SAR images account for approximately 90\% of the SAR images in VesselBench-800K.}

\subsection{Data Annotations}
As shown in Table \ref{tab:dataacquisition}, the images from the six data sources require labeling. Specifically, the DF contests and DOTA datasets need to be manually relabeled to meet the requirement for HBB annotations, as well as to address missing instances (e.g., small or densely clustered vessels) or unlabeled instances (e.g., those from the testing set in the contests). Furthermore, the remaining four sources are entirely unlabeled in terms of vessels. We accordingly employ the LabelMe tool \cite{russell2008labelme} for manual labeling of all these images.

{\textbf{Annotation Quality Control.} All vessel annotations in VesselBench-800K are manually generated or manually re-checked using LabelMe. For existing datasets with vessel annotations, we manually inspect and revise the original labels to maintain consistent HBB annotation criteria, particularly for missing or densely distributed small vessels. For unlabeled images, vessel instances are manually annotated and subsequently checked to reduce annotation errors. Particular attention is paid to challenging cases, including tiny vessels, vessel wakes, and cluttered coastal or harbor backgrounds, which are the major sources of annotation ambiguity.}

\subsection{Data Characteristics}
The data in our VesselBench-800K exhibit several significant characteristics: 1) diverse types, sizes, and densities of vessels; 2) rich scenes and backgrounds, ranging from simple to complex covers; 3) a broader range of image spatial (high) resolution, i.e., 0.1$\sim$4.5m;  4) a massive dataset of 800,000 multimodal images; 5) multi-scale vessels appearing within the same and different scenes; 6) extensive geographical coverage on a global scale. Fig. \ref{fig:samples} illustrates some typical examples from VesselBench-800K to showcase key data characteristics.

\begin{figure*}[!t]
      \centering
	   \includegraphics[width=1.0\textwidth]{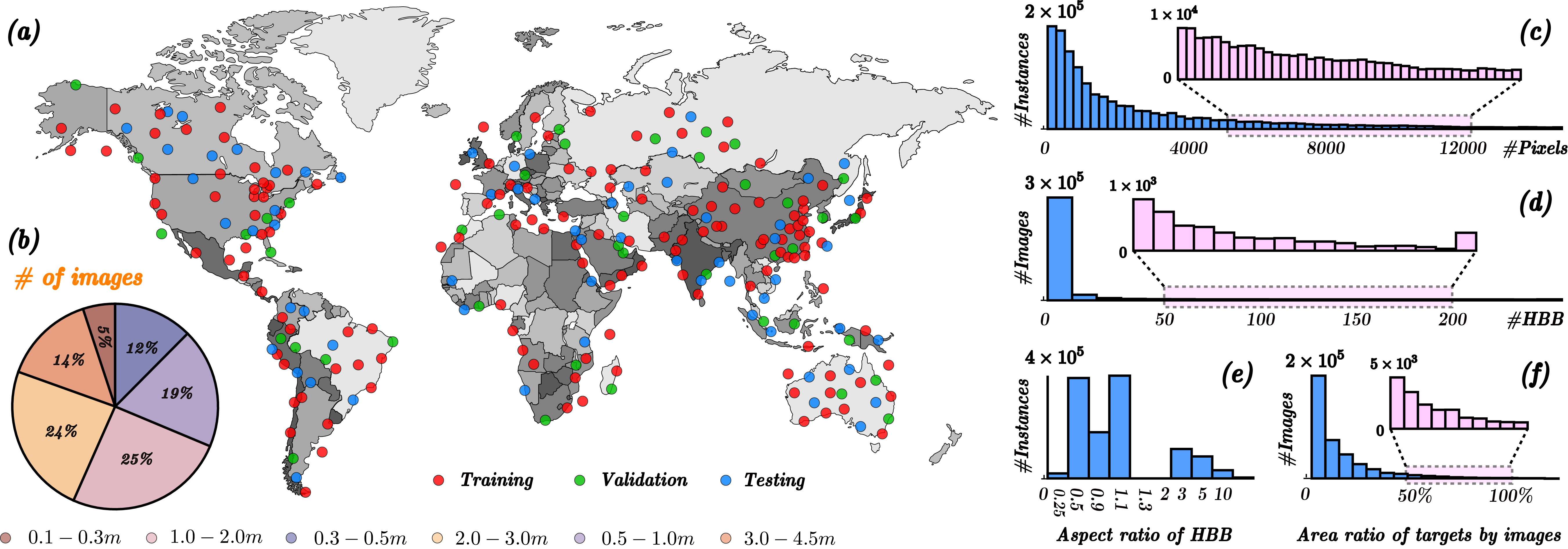}
      \caption{An illustrative data statistics of VesselBench-800K. (a) Spatial distribution of data collection on a global scale, categorized into training, validation, and testing sets in \textbf{{red}}, \textbf{\textcolor{green}{green}}, and \textbf{\textcolor{blue}{blue}}, respectively. (b) gives the number of images across different resolutions (GSD). (c) shows pixel-level statistics of vessel instances. (d) collects the number of HBB per image. (e) counts the number of instances in different aspect ratios of HBB. (f) calculates the area ratio of targets within images.}
\label{fig:distribution}
\end{figure*} 

\subsection{Data Statistics}
There is no doubt that the proposed VesselBench-800K is a large-scale vessel perception benchmark dataset for multimodal RS images, with significant potential for applications in global maritime activities. Fig. \ref{fig:distribution} shows an illustrative statistical overview of VesselBench-800K. To highlight its extensive geographical coverage, we visualize a rough spatial distribution of data collection on a global scale, as shown in Fig. \ref{fig:distribution}(a), while Fig. \ref{fig:distribution}(b) gives a proportional result of the number of images across different GSDs. Moreover, Figs. \ref{fig:distribution}(c)-(f) illustrate various dataset statistics in terms of the number of instances across different pixel counts, the number of images based on the count of HBB, the distribution of instances with different aspect ratios of HBB, and the number of images with different target area ratios, respectively.

{These statistics also reveal substantial variations and imbalanced distributions in vessel scale, density, aspect ratio, and target occupancy, reflecting important long-tail characteristics of real-world maritime vessel perception. Since the current benchmark adopts a unified vessel category, fine-grained vessel-type imbalance is not explicitly quantified in this work.}

\begin{table*}[!t]
    \centering
    \caption{Quantitative performance comparison of vessel detection methods using unimodal and multimodal RS images on VesselBench-800K, where eight metrics, including mAP, AP$_{50}$, AP$_{75}$, AP$_{s}$, AP$_{m}$, AP$_{l}$, as well as FLOPs and parameters of networks, are compared. FPN refers to feature pyramid networks \cite{lin2017feature} and CM denotes ChannelMapper \cite{sylvester2021autogenic}. The best results for different modalities are highlighted in bold, respectively.}
    \resizebox{1\textwidth}{!}{
    \begin{tabular}{c|c|c|ccc|ccc|cc}
        \toprule[1.5pt]
        Methods & Backbone & Epochs & AP & AP$_{50}$ & AP$_{75}$ & AP$_{s}$ & AP$_{m}$ & AP$_{l}$ & Parameters & FLOPs\\
        \hline
        \addlinespace[2pt]
        \multicolumn{11}{c}{Unimodal \textbf{\textcolor{orange}{Optical}} vessel detection}\\
        \addlinespace[2pt]
        \hline
        \addlinespace[1.5pt]
        Faster R-CNN \cite{ren2016faster} & ResNet50-FPN & 36 & 0.512 & 0.761 & 0.568 & 0.400 & 0.731 & 0.707 & 41.348M & 0.148T\\
        RetinaNet \cite{ross2017focal} & ResNet50-FPN & 36 & 0.536 & 0.762 & 0.573 & 0.402 & 0.750 & 0.752 & 36.351M & 0.128T\\
        Mask R-CNN \cite{he2017mask} & ResNet50-FPN & 36 & 0.552 & 0.736 & 0.617 & 0.405 & 0.785 & \textbf{0.831} & 43.971M & 0.276T\\
        AutoAssign \cite{zhu2020autoassign} & ResNet50-FPN & 36 & 0.532 & 0.752 & 0.591 & 0.399 & 0.742 & 0.712 & 36.426M & 0.126T\\
        TOOD \cite{feng2021tood} & ResNet50-FPN & 36 & \textbf{0.586} & \textbf{0.775} & \textbf{0.641} & \textbf{0.445} & \textbf{0.807} & 0.822 & \textbf{32.018M} & 0.123T\\
        DINO \cite{zhang2022dino} & ResNet50-CM & 36 & 0.479 & 0.743 & 0.502 & 0.289 & 0.775 & 0.867 & 47.465M & \textbf{0.024T}\\
        \hline
        \addlinespace[2pt]
        \multicolumn{11}{c}{Unimodal \textbf{\textcolor{darkred}{SAR}} vessel detection}\\
        \addlinespace[2pt]
        \hline
        \addlinespace[1.5pt]
        Faster R-CNN \cite{ren2016faster} & ResNet50-FPN & 36 & 0.432 & 0.681 & 0.466 & 0.310 & 0.658 & 0.651 & 41.348M & 0.148T\\
        RetinaNet \cite{ross2017focal} & ResNet50-FPN & 36 & 0.466 & 0.698 & 0.492 & 0.321 & 0.693 & 0.725 & 36.351M & 0.128T\\
        Mask R-CNN \cite{he2017mask} & ResNet50-FPN & 36 & 0.483 & 0.686 & 0.541 & 0.329 & 0.725 & \textbf{0.786} & 43.971M & 0.276T\\
        AutoAssign \cite{zhu2020autoassign} & ResNet50-FPN & 36 & 0.495 & \textbf{0.699} & 0.541 & 0.342 & 0.726 & 0.770 & 36.426M & 0.126T\\
        TOOD \cite{feng2021tood} & ResNet50-FPN & 36 & \textbf{0.501} & 0.696 & \textbf{0.545} & \textbf{0.353} & \textbf{0.736} & 0.785 & \textbf{32.018M} & 0.123T\\
        DINO \cite{zhang2022dino} & ResNet50-CM & 36 & 0.416 & 0.674 & 0.428 & 0.223 & 0.709 & 0.839 & 47.465M & \textbf{0.024T}\\
        \hline
        \addlinespace[2pt]
        \multicolumn{11}{c}{\textbf{\textcolor{custompurple}{Multimodal}} vessel detection}\\
        \addlinespace[2pt]
        \hline
        \addlinespace[1.5pt]
        Faster R-CNN \cite{ren2016faster} & ResNet50-FPN & 36 & 0.560 & 0.790 & 0.624 & 0.447 & 0.770 & 0.787 & 41.413M & 0.157T\\
        RetinaNet \cite{ross2017focal} & ResNet50-FPN & 36 & 0.521 & 0.746 & 0.557 & 0.385 & 0.735 & 0.740 & 36.416M & 0.136T\\
        Mask R-CNN \cite{he2017mask} & ResNet50-FPN & 36 & 0.554 & 0.729 & 0.612 & 0.407 & 0.787 & 0.834 & 44.036M & 0.132T\\
        AutoAssign \cite{zhu2020autoassign} & ResNet50-FPN & 36 & 0.594 & 0.779 & 0.648 & 0.457 & 0.803 & 0.821 & 36.309M & 0.132T\\
        TOOD \cite{feng2021tood} & ResNet50-FPN & 36 & \textbf{0.629} & \textbf{0.805} & \textbf{0.676} & \textbf{0.489} & \textbf{0.838} & \textbf{0.878} & \textbf{32.083M} & 0.132T\\
        DINO \cite{zhang2022dino} & ResNet50-CM & 36 & 0.578 & 0.798 & 0.623 & 0.416 & 0.816 & 0.877 & 47.530M & \textbf{0.025T}\\
        \bottomrule[1.5pt]
    \end{tabular}
    }
    \label{tab:detection}
\end{table*}

\begin{figure*}[!t]
      \centering
	   \includegraphics[width=1.0\textwidth]{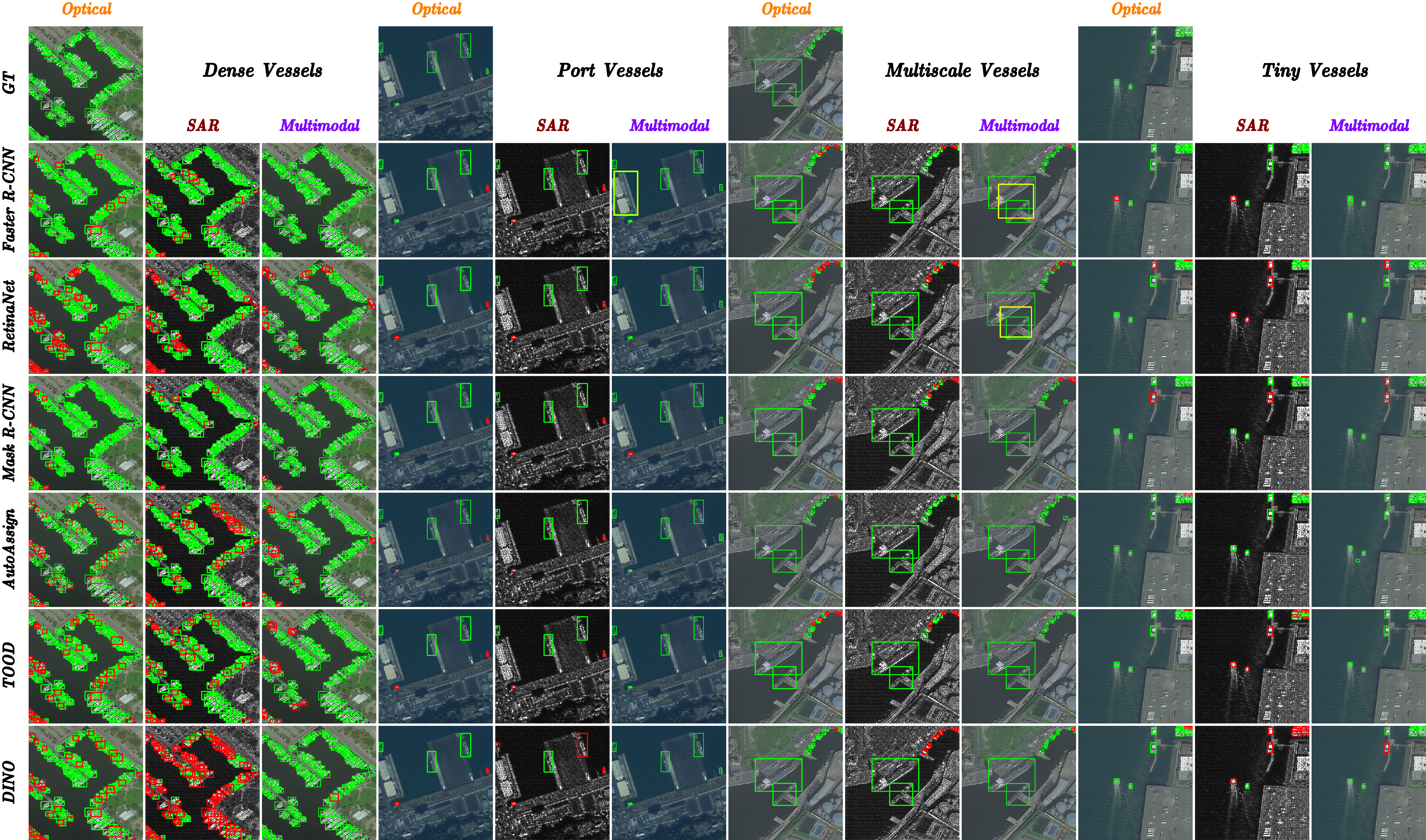}
      \caption{Visualization of exemplar results from different vessel detection methods in unimodal and multimodal scenarios on VesselBench-800K. The HBBs in green, red, and yellow denote true positives, false negatives, and false positives, respectively.}
\label{fig:Det}
\end{figure*} 

\section{Evaluations on VesselBench-800K}
\subsection{Implementation Details}
We randomly partition the VesselBench-800K dataset into training, validation, and testing sets in a 7:1:2 ratio, selecting samples from the entire set of original images. Furthermore, the images with GTs for the training and validation sets are openly and publicly available, while the testing set remains private to support the development of an evaluation platform. Fig. \ref{fig:distribution}(a) visualizes the spatial distributions of the training, validation, and testing sets across global geographical coverage. {The training and testing sets are randomly divided from the overall dataset to provide a unified benchmark for different vessel perception methods. We note that such a random split may not completely exclude spatial or temporal correlations between samples, particularly for images acquired from nearby regions or under similar conditions.}

In the vessel detection task, all methods are implemented on the MMdetection platform, using an Intel$^{\circledR}$ Xeon$^{\textregistered}$ Platinum 8383C CPU @ 2.70GHz $\times$ 160 processors, 512GB RAM, and a 24GB NVIDIA GTX4090 GPU for training. All models use ResNet-50 as the backbone and are trained for 36 epochs with a batch size of 8.  Apart from DINO, which combines ResNet-50 and ChannelMapper (CM) \cite{sylvester2021autogenic}, the remaining methods adapt Feature Pyramid Networks (FPN) \cite{lin2017feature} with ResNet-50. The optimizer and initial learning rate are set according to the default settings of each method. For Faster R-CNN, TOOD, AutoAssign, and RetinaNet, SGD is used as the optimizer, while the remaining methods use AdamW \cite{kingma2014adam} with a weight decay of $1\times10^{-5}$. 

For vessel counting and density estimation, we select images containing vessel targets from the entire set of original images partitioned for the detection task to form the training, validation, and testing sets. All experiments are implemented on the PyTorch framework with a batch size of 4, and the remaining hyperparameters are set according to the default configurations of each method.

\subsection{Evaluation Metrics}
{VesselBench-800K provides standardized evaluation protocols for the benchmark tasks with different forms of ground-truth labels. For vessel detection, HBB annotations are used as ground truth for localizing individual vessels. For vessel counting and density estimation, the vessel counts and corresponding density maps derived from the annotated vessel instances are used as ground truth. All tasks follow the same predefined training, validation, and testing splits to ensure consistent and reproducible evaluation.}

We evaluate the vessel detection performance both qualitatively and quantitatively using six commonly used indices: AP, AP$_{50}$, AP$_{75}$, AP$_{s}$, AP$_{m}$, AP$_{l}$. Since our task only considers one vessel category, AP is calculated at 10 different IoU thresholds, i.e., IoU@[0.5:0.05:0.95] and then averaged. AP50 and AP75 denote the AP values at IoU thresholds of 50 and 75, respectively. In addition to calculating AP based on different IoU thresholds, it can also be computed based on the size of the detected objects. APs, APm, and APl correspond to the AP values for objects with areas $\leq$ 32$\times$32 pixels, 32$\times$32 pixels $\textless$ area $\leq$ 96$\times$96 pixels, and area $\textgreater$ 96$\times$96 pixels, respectively. We adopt MAE, MSE, PSNR, and SSIM as the evaluation metrics for the counting and density estimation tasks. {The scale-specific metrics APs, APm, and APl further provide a stratified evaluation of detection performance for vessels of different sizes, which is particularly relevant to the substantial scale variations in VesselBench-800K.}

\subsection{SOTA methods for Comparisons}
We select several well-known and SOTA vessel detection methods for comparisons on VesselBench-800K, including Faster R-CNN \cite{ren2016faster}, RetinaNet \cite{ross2017focal}, Mask R-CNN \cite{he2017mask}, AutoAssign \cite{zhu2020autoassign}, TOOD \cite{feng2021tood}, and DINO \cite{zhang2022dino}. The detection performance of these methods, using HBB, is evaluated in both unimodal and multimodal scenarios. Note that we adopt the same network backbone architectures for unimodal and multimodal vessel detections. The main difference lies in the number of modalities and dimensions in the network input. Unlike the unimodal case, where data are directly fed into the network, the multimodal architecture begins with a two-stream CNN feature extractor to process optical and SAR data, respectively, before passing the extracted features further into the main backbone of the network. {This unified multimodal setting is adopted to provide a consistent baseline for evaluating the contribution of optical-SAR observations across different detection methods. Since the primary objective of this work is to establish a large-scale multimodal vessel perception dataset and benchmark, rather than to exhaustively investigate different multimodal fusion architectures, more sophisticated fusion strategies are not systematically compared in the current study.}

For the vessel counting and density estimation task, we evaluate the performance of five SOTA methods on VesselBench-800K. These methods include MCNN \cite{zhang2016single}, CSRNet \cite{li2018csrnet}, SCAR \cite{gao2019scar}, SFCN+ \cite{wang2019learning}, and PET \cite{liu2023point}. The first four methods are mainly dominated by density estimation, while the last one places more emphasis on counting.

\subsection{Vessel Detection}
Fig. \ref{fig:Det} shows the vessel detection results of several SOTA methods visually in both unimodal and multimodal scenarios using the proposed VesselBench-800K datasets. Correspondingly, Table \ref{tab:detection} qualifies the vessel detection performance among these compared models using unimodal and multimodal RS images on VesselBench-800K, in terms of AP, AP$_{50}$, AP$_{75}$, AP$_{s}$, AP$_{m}$, AP$_{l}$, as well as FLOPs and parameters of networks.

By and large, the ability to detect vessels on SAR images remains limited compared to optical images, particularly for tiny and dense vessels, which are often not detected accurately. The detection performance using multimodal RS images is superior to that of only using single modalities, such as optical or SAR alone.  However, it should be noted that some false positives are generated when using multimodal images. This indicates that overly incorporating information may hurt the detection system, leading to false positive detections. Furthermore, TOOD achieves the best result for multimodal vessel detection with the fewest parameters among all the compared methods. Although there is a slight fluctuation when applied to unimodal SAR, TOOD still holds an advantage in unimodal image detection.

{It is also worth noting that a domain gap remains between real and synthetic SAR images due to differences in imaging mechanisms and data distributions. Meanwhile, variations across sensors, spatial resolutions, modalities, and maritime scenes further challenge the cross-domain robustness of current models. Typical failure cases are mainly observed for tiny and densely distributed vessels, as well as complex backgrounds that may lead to false-positive detections. These observations indicate that improving real-to-synthetic domain generalization and cross-domain robustness remains an important direction for future vessel perception.}

\begin{figure*}[!t]
      \centering
	   \includegraphics[width=0.7\textwidth]{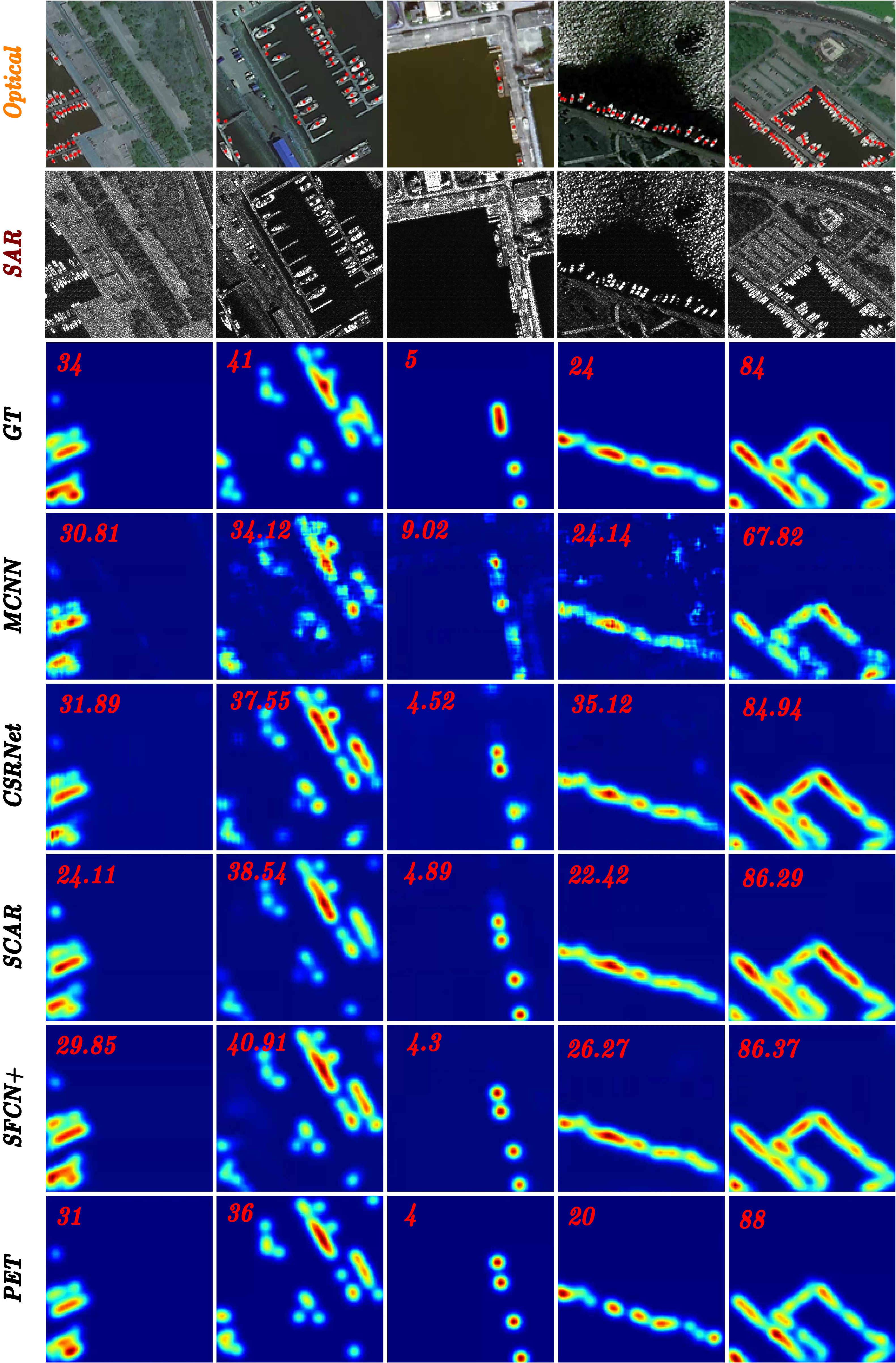}
      \caption{Visualization of different compared methods for vessel counting and density estimation in the selected five scenes of VesselBench-800K. Note that the heat maps reflect the density estimation for vessels, while the numbers in red denote the GT and estimated results of vessel counting.}
\label{fig:CDE}
\end{figure*} 

\begin{table}[!t]
    \centering
    \caption{Quantitative results of different compared methods for vessel density estimation using multimodal RS images on VesselBench-800K, where four main indices, i.e., MAE, MSE, PSNR, and SSIM, are computed. The best results are highlighted in bold.}
    \resizebox{0.47\textwidth}{!}{
    \begin{tabular}{c|c|c|c|c|c}
        \toprule[1.5pt]
        Methods & Epochs & MAE & MSE & PSNR & SSIM\\
        \hline
        \addlinespace[1.5pt]
        MCNN \cite{zhang2016single} & 27 & 2.714 & 35.468 & 23.088 & 0.764\\
        CSRNet \cite{li2018csrnet} & 26 & 1.524 & 20.277 & 29.382 & 0.926\\
        SCAR \cite{gao2019scar} & 18 & \textbf{1.417} & 20.860 & \textbf{31.610} & \textbf{0.958}\\
        SFCN+ \cite{wang2019learning} & 26 & 1.462 & 19.726 & 31.193 & 0.945\\
        PET \cite{liu2023point} & 200 & 1.629 & \textbf{3.98} & -- & --\\
        \bottomrule[1.5pt]
    \end{tabular}
    }
    \label{tab:estimation}
\end{table}

\subsection{Vessel Counting and Density Estimation}
Fig. \ref{fig:CDE} and Table \ref{tab:estimation} give quantitative and qualitative results regarding vessel counting and density estimation, respectively. Overall, SCAR achieves the best performance in MAE, PSNR, and SSIM with fewer epochs, while PET excels in MSE, owing to its stronger focus on counting. This, to a great extent, demonstrates the effectiveness of the attention mechanism in both vessel counting and density estimation tasks. Visually, the estimated results of density maps closely align with the quantitative findings, with SCAR producing results that most closely match the GTs. Vessel number estimation remains challenging, as the results show a relatively large difference between GTs and estimated values, despite PET being specifically designed for counting tasks.

\section{Conclusions}
We introduce VesselBench-800K, a large-scale dataset for vessel perception in multimodal RS images. VesselBench-800K consists of 800,000 optical and SAR images and 1,362,662 instances at GSDs of 4.5m to 0.1m, capturing a diverse array of vessel types, sizes, and densities. To the best of our knowledge, VesselBench-800K is the largest dataset to date for vessel perception, globally collected and integrated from various platforms, sensors, scenes, backgrounds, existing datasets, and synthetic sources. 

More notably, this is the first multimodal heterogeneous (optical and SAR) benchmark, enabling all-weather, all-day maritime traffic monitoring and perception. In addition, we manually annotate a vast number of vessel samples with HBB and generate the corresponding density maps to ensure high-quality ground truth labels. We also evaluate the perception performance beyond general target detection, allowing higher-level perception tasks on VesselBench-800K, such as vessel counting and density estimation.

\vspace{2pt}
\noindent \textbf{Limitations and future insights.} In the vessel detection task, we focus solely on the HBB case. However, oriented targets, particularly vessels in real-world offshore operations, are common. This limitation may result in performance degradation, as HBBs do not adequately account for the orientation of vessels. 

{In future work, we plan to extend the current HBB annotations to oriented bounding boxes (OBBs) and release them in the next version of VesselBench-800K, enabling more accurate detection of vessels with arbitrary orientations, particularly in dense and complex maritime environments. Additionally, finer vessel recognition and classification will be explored through instance segmentation and fine-grained recognition to further enhance vessel perception capabilities. Beyond these task-specific extensions, VesselBench-800K can serve as a large-scale benchmark for future research on multimodal foundation models, cross-modal learning between optical and SAR observations, and its extension to additional modalities, as well as large-scale maritime intelligence applications. More comprehensive studies of multimodal fusion strategies, including early, late, and attention-based fusion, will also be investigated based on VesselBench-800K in our subsequent work. In addition, fine-grained annotations of vessel types, rare events, and environmental conditions (e.g., weather, illumination, and sea state) are not consistently available across all constituent data sources, which currently limits rigorous analyses of long-tail distributions and type- or condition-specific performance. We will further enrich these annotations to enable more comprehensive and stratified evaluations under diverse maritime scenarios. With more complete environmental annotations, future versions of VesselBench-800K will further support systematic evaluation of multimodal perception under different maritime conditions.}

{Another limitation lies in the current data-splitting protocol. VesselBench-800K integrates multiple existing public datasets together with newly collected and generated data, for which complete and consistent acquisition time information is not available. Therefore, strict cross-region and cross-time evaluation protocols are not included in the current version. In future versions, we will further organize the spatial and temporal metadata and establish dedicated cross-region and cross-time splits to better evaluate the generalization and robustness of vessel perception models.}

\section*{Acknowledgements}
Our VesselBench-800K dataset is constructed from three primary sources: (1) publicly available open-source datasets, (2) satellite imagery obtained through manually accessible platforms (e.g., Google Earth), and (3) synthetic data generated using generative models (e.g., CycleGAN). Each open-source dataset included in VesselBench-800K was carefully verified to ensure compliance with its licensing terms, which permit use for non-commercial, academic research. More specifically:

DOTA \cite{xia2018dota}: All images and associated annotations are released for academic purposes only; commercial use is prohibited. Images originating from Google Earth remain subject to the Google Earth Terms of Service.

GLH-water \cite{li2024glh}: A publicly released large-scale surface water detection dataset derived from very high-resolution satellite imagery, distributed under an open-access academic research policy.

GLH-Bridge \cite{li2024learning}: Released by the School of Remote Sensing and Information Engineering, Wuhan University, with terms restricting use to academic research only; all commercial use is strictly prohibited. Google Earth imagery contained within the dataset is further subject to the Google Earth Terms of Service.

QXS-SAROPT \cite{huang2021qxs}: A publicly available dataset for optical and radar ship-target tasks, provided exclusively for non-commercial, academic research.

In addition, supplementary images were collected directly from the Google Earth platform for academic research purposes. Such use fully complies with the official \href{https://about.google/brand-resource-center/products-and-services/geo-guidelines/}{ ``Google Earth'' terms of use}, with proper attribution as: ``Google Earth images $\copyright$ Google. All rights reserved.''

We sincerely thank all participants who contributed to the data download and annotations of VesselBench-800K.

\bibliographystyle{IEEEtran}
\bibliography{ref}

\end{document}